\documentclass[10pt,twocolumn,letterpaper]{article}

\usepackage{cvpr}
\usepackage{times}
\usepackage{epsfig}
\usepackage{graphicx}
\usepackage{amsmath}
\usepackage{amssymb}
\usepackage{pifont}

\usepackage{float}
\usepackage{lipsum}
\usepackage{stfloats}
\usepackage{multicol}
\usepackage{multirow}
\usepackage{amsmath,bm}
\usepackage{etoolbox}

 \usepackage[export]{adjustbox}
 \usepackage{algorithm}
 \usepackage{mathtools}
 \usepackage{algorithmic}
 \usepackage{wrapfig,booktabs}
 \usepackage{wrapfig}
 \usepackage{array}

\usepackage{array}

\usepackage{tabulary}

\usepackage{paralist}

\usepackage{graphicx}
\usepackage{color}
\usepackage[dvipsnames]{xcolor}
\usepackage{booktabs}
\usepackage{mathtools}
\usepackage[export]{adjustbox}

\usepackage[caption=false,font=footnotesize]{subfig}

\usepackage[pagebackref=true,breaklinks=true,letterpaper=true,colorlinks,bookmarks=false]{hyperref}
\hypersetup{colorlinks, citecolor=teal}
\hypersetup{colorlinks, linkcolor=teal}

\cvprfinalcopy 

\def\cvprPaperID{8337} 

\ifcvprfinal\fi
\begin{document}

\title{Temporally-Weighted Hierarchical Clustering for Unsupervised Action Segmentation}

\author{M. Saquib Sarfraz$^{1,7}$,  Naila Murray$^{2}$, Vivek Sharma$^{1,3,4}$, Ali Diba$^{5}$,  Luc Van Gool$^{5,6}$, Rainer Stiefelhagen$^{1}$ \\
\small{$^{1}$ Karlsruhe Institute of Technology,
$^{2}$ Facebook AI Research,}\\
\small{$^{3}$ MIT,
$^{4}$ Harvard Medical School,
$^{5}$ KU Leuven,
$^{6}$ ETH Zurich
$^{7}$ Daimler TSS}
}
\maketitle

\begin{abstract}
 Action segmentation refers to inferring boundaries of semantically consistent visual concepts in videos and is an important requirement for many video understanding tasks. 
 For this and other video understanding tasks, supervised approaches have achieved encouraging performance but require a high volume of detailed frame-level annotations. We present a fully automatic and unsupervised approach for segmenting actions in a video that does not require any training. Our proposal is an effective temporally-weighted hierarchical clustering algorithm that can group semantically consistent frames of the video. Our main finding is that representing a video with a 1-nearest neighbor graph by taking into account the time progression is sufficient to form semantically and temporally consistent clusters of frames where each cluster may represent some action in the video. Additionally, we establish strong unsupervised baselines for action segmentation and show significant performance improvements over published unsupervised methods on five challenging action segmentation datasets.
 Our code is available.\footnote{\text{https://github.com/ssarfraz/FINCH-Clustering/tree/master/TW-FINCH}}
 

\end{abstract}
\section{Introduction}
\begin{figure*}[hbt]
\centering
\includegraphics[width=2\columnwidth]{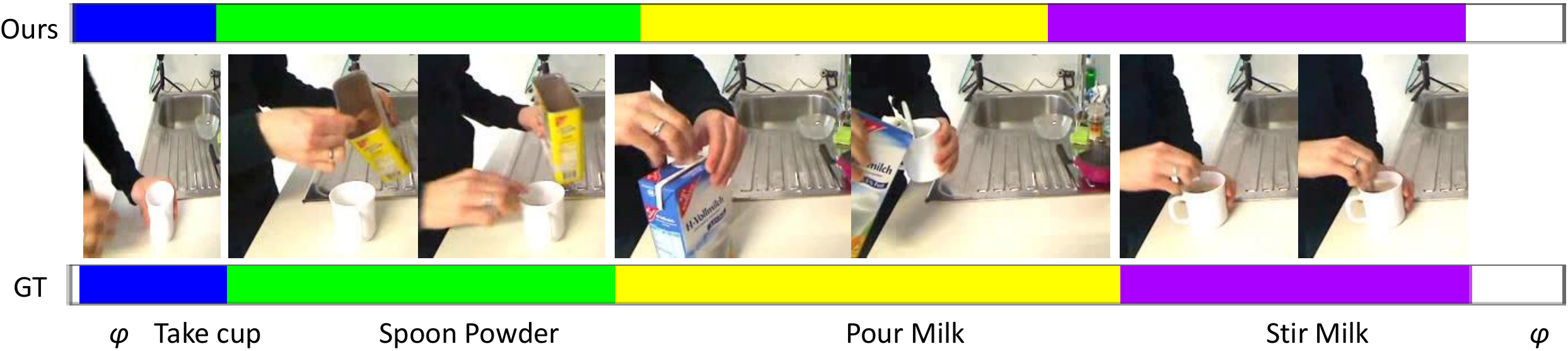}

\caption{Segmentation output example from Breakfast Dataset~\cite{ute_15}: \textit{P46\_webcam02\_P46\_milk}.  Colors indicate different actions in chronological order:  $\varphi$, take\_cup, spoon\_powder, pour\_milk, stir\_milk, $\varphi$, where $\varphi$ is background shown in while color.\label{fig:teaser} }
\vspace{-0.3cm}
\end{figure*}

Human behaviour understanding in videos has traditionally been addressed by inferring high-level semantics such as activity recognition \cite{herath2017going,asadi2017deep}.
Such works are often limited to tightly clipped video sequences to reduce the level of labelling ambiguity and thus make the problem more tractable.
However, a more fine-grained understanding of video content, including for un-curated content that may be untrimmed and therefore contain a lot of material unrelated to human activities, would be beneficial for many downstream video understanding applications.
Consequently, the less-constrained problem of action segmentation in untrimmed videos has received increasing attention.
Action segmentation refers to labelling each frame of a video with an action, where the sequence of actions is usually performed by a human engaged in a high-level activity such as making coffee (illustrated in Figure~\ref{fig:teaser}).
Action segmentation is more challenging than activity recognition of trimmed videos for several reasons, including the presence of background frames that don't depict actions of relevance to the high-level activity.
A major challenge is the need for significantly more detailed annotations for supervising learning-based approaches.
For this reason, weakly- and unsupervised approaches to action segmentation have gained popularity \cite{mucon,actionset,ute_paper,mallow}.
Some approaches have relied on natural language text extracted from accompanying audio to provide frame-based action labels for training action segmentation models \cite{yti_paper}. This of course makes the strong assumption that audio and video frames are well-aligned. Other approaches assume some \textit{a priori} knowledge of the actions, such as the high-level activity label or the list of actions depicted, in each video \cite{sct,mucon}. Even this level of annotation however, requires significant annotation effort for each training video as not all activities are performed using the same constituent actions. 

Most weakly- and unsupervised methods, whatever their degree of \textit{a priori} knowledge, focus on acquiring pseudo-labels that can be used to supervise training of task-specific feature embeddings \cite{cdfl,mallow,ute_7,ute_24,nnv,mucon}. As pseudo-labels are often quite noisy, their use may hamper the efficacy of the learned embeddings. In this work, we adopt the view that action segmentation is fundamentally a $grouping$ problem, and instead focus on developing clustering methods that effectively delineate the temporal boundaries between actions. This approach leads to an illuminating finding: for action segmentation, a simple clustering ($e.g.$, with Kmeans) of appearance-based frame features achieves performance on par with, and in some cases superior to, SoTA weakly-supervised and unsupervised methods that require training on the target video data (please refer to section~\ref{sec:method} for details).
This finding indicates that a sufficiently discriminative visual representation of video frames can be used to group frames into visually coherent clusters.
However, for action segmentation, temporal coherence is also critically important.
Building on these insights, we adapt a hierarchical graph-based clustering algorithm to the task of temporal video segmentation by modulating appearance-based graph edges between frames by their temporal distances. The resulting spatio-temporal graph captures both visually- and temporally-consistent neighbourhoods of frames that can be effectively extracted.
Our work makes the following main contributions:
\begin{itemize}
    \item We establish strong appearance-based clustering baselines for unsupervised action segmentation that outperform SoTA models;
    \item We propose to use temporally-modulated appearance-based graphs to represent untrimmed videos;
    \item We combine this representation with a hierarchical graph-based clustering algorithm in order to perform temporal action segmentation.
\end{itemize}
Our proposed method outperforms our strong baselines and existing SOTA unsupervised methods by a significant margin on 5 varied and challenging benchmark datasets.

\section{Related Work}
There exists a large body of work on spatial and spatio-temporal action recognition in videos (see \cite{herath2017going,asadi2017deep} for recent surveys). In this section we review works related to our problem of interest, temporal action segmentation, focusing on weakly- and unsupervised methods.

Most existing temporal action segmentation methods, be they fully supervised~\cite{ute_6,ute_15,kuehne2018hybrid}, weakly supervised~\cite{mucon,cdfl,nnv,sct} or unsupervised~\cite{ute_paper,mallow,yti_paper}, use frame-level annotations to train their models. They differ in whether the annotations are collected by human annotators or extracted in a semi- or unsupervised manner. 
These models largely follow a paradigm in which an embedding is trained on top of pre-extracted frame-level video features, such as I3D~\cite{i3d}, as in~\cite{mucon,sct}, or hand-crafted video features such as improved dense trajectories IDT~\cite{idt}, as in~ \cite{cdfl,mallow,ute_paper,ute_7,ute_24}.
To train the embedding, a discriminative objective function is used in conjunction with the collected annotations~\cite{cdfl,mallow,ute_7,ute_24,nnv,mucon}. Weakly-supervised and unsupervised methods, discussed next, vary largely in the manner in which they extract and exploit pseudo-labels.
 
\noindent\textbf{Weakly-supervised methods} generally assume that both the video-level activity label and the ordering of actions, termed transcripts, are known during training.
Some weakly-supervised works have a two-stage training process where pseudo-labels are first generated using transcripts and then used to train a frame classification network~\cite{ute_17,ute_24}.
In contrast, the method NN-Vit~\cite{nnv} directly leverages transcripts while learning a frame classification model. For this they introduce a loss based on Viterbi decoding that enforces consistency between frame-level label predictions. 
In a similar spirit, a recent proposal called MuCoN~\cite{mucon} aims to leverage transcripts while learning a frame classification model. They learn two network branches, only one of which has access to transcripts, while ensuring that both branches are mutually consistent.
Another recent method called CDFL~\cite{cdfl} also aims to use transcripts when training their frame labelling model. They first build a fully-connected, directed segmentation graph whose paths represent actions. They then train their model by maximizing the energy difference between valid paths ($i.e$ paths that are consistent with the ground-truth transcript) and invalid ones. 
In SCT~\cite{sct}, the authors assume that the set of action labels for a given video, but not their order, is known. They determine the ordering and temporal boundaries of the actions by alternatively optimizing set and frame classification objectives to ensure that frame-level action predictions are consistent with the set-level predictions.

\noindent\textbf{Unsupervised methods} generally assume knowledge only of the video-level activity label~\cite{mallow, yti_paper, ute_paper, lstm_al,vt_unet}.
In Mallow~\cite{mallow}, the authors use video-level annotations in an iterative approach to action segmentation, alternating optimization of a discriminative appearance model and a generative temporal model of action sequences.
In Frank-Wolfe~\cite{yti_paper}, video narrations are extracted using ASR and used to extract an action sequence for a set of videos of an activity. This is accomplished by separately clustering the videos and the ASR-recovered speech to identify action verbs in the specific video. Temporal localization is then obtained by training a linear classifier. 
CTE~\cite{ute_paper} proposes to learn frame embeddings that incorporate relative temporal information. They train a video activity model using pseudo-labels generated from Kmeans clustering of the videos' IDT features. The trained embeddings are then re-clustered at the ground-truth number of actions and ordered using statistics of the relative time-stamps with a GMM+Viterbi decoding. VTE-UNET~\cite{vt_unet} uses similarly learned embeddings in combination with temporal embeddings to improve upon \cite{ute_paper}. Another interesting approach is
LSTM+AL~\cite{lstm_al}, which fine-tunes a pre-trained VGG16 model with an LSTM, using future frame prediction as a self-supervision objective, to learn frame embeddings. These embeddings are then used to train an action boundary detection model.

All of these methods require training on the target video dataset, which from a practical standpoint is a very restrictive requirement. 
In contrast,
our method does not require any training, and relies only on frame clustering to segment a given video.
\section{Method} \label{sec:method}
As mentioned in the introduction, unsupervised temporal video segmentation is inherently a grouping and/or clustering problem. We observe that, given a relatively good video frame representation, the boundaries of actions in a video are discernible without the need for further training on objectives that use noisy pseudo-labels, something that almost all current methods pursue. To substantiate this observation and to have a basis for our later discussion we provide results of directly clustering a commonly used untrimmed video benchmark (Breakfast dataset~\cite{ute_15} with 1712 videos) in Table~\ref{table:baseclustering}.
The goal of clustering is to group the frames of each video into its ground-truth actions.
We consider two representative clustering methods: (1) Kmeans~\cite{kmeans}, representing centroid-based methods; and (2) a recent proposal called FINCH~\cite{finch}, representing state-of-the-art hierarchical agglomerative clustering methods. We cluster the extracted 64-dim IDT features of each video to its required number of actions (clusters). The performance is computed by mapping the estimated cluster labels of each video to the ground-truth labels using the Hungarian method, and the accuracy is reported as mean over frames (MoF). Section~\ref{sec:exp} contains more details about the experimental setup.
As can be seen, simple clustering baselines Kmeans/FINCH performs at par with the best reported weakly/un-supervised methods in this video level evaluation.
These results establish new, strong baselines for temporal video segmentation, and suggest that focusing on more specialized clustering techniques may be promising.

Among existing clustering methods, hierarchical clustering methods such as~\cite{finch} are an attractive choice for the task at hand, as such methods provide a hierarchy of partitions of the data as opposed to a single partition. In this paper we adopt a hierarchical clustering approach to action segmentation that does not require video-level activity labels.
In contrast, the existing body of work requires not only such prior knowledge but also requires training on the target video data.
The ability to generate a plausible video segmentation without relying on training is highly desirable from a practical standpoint.
To the best of our knowledge there is no existing prior work that addresses this challenging and practical scenario.
\begin{table}[t!]
\centering
\resizebox{8.3cm}{!}{
\begin{tabular}{c|c|c|c||c|c}
\toprule
&{\color{blue}{Weakly Sup.}} &\multicolumn{4}{c}{\color{blue}{Unsupervised}}\\
&CDFL~\cite{cdfl} & LSTM+AL~\cite{lstm_al} & VTE-UNET~\cite{vt_unet} & Kmeans & FINCH 
\\
\midrule
MoF & 50.2 &42.9 &52.2 & 42.7 & 51.9 
\\
\bottomrule
\end{tabular}}
\caption{Simple clustering with Kmeans or FINCH is competitive with the best reported weakly or unsupervised methods. \label{table:baseclustering}}
\vspace{-0.3cm}
\end{table}

Our proposal is similar in spirit to the FINCH~\cite{finch} algorithm. The authors in~\cite{finch} make use of the observation that the nearest and the shared neighbor of each sample can form large linking chains in the data. They define an adjacency matrix that links all samples to their nearest first neighbour, thus building a 1-nearest neighbor (1-NN) graph. They showed that the connected components of this adjacency graph partitions the data into fixed clusters. A recursive application of this on the obtained partition(s) yields a hierarchy of partitions. The algorithm typically provides hierarchical partitions of the data in only a few recursion steps.

Based on this observation of finding linking chains in the data with nearest or shared neighbours, we propose a similar hierarchical clustering procedure for the problem of temporal video segmentation. 
We propose to use a spatio-temporal graphical video representation by linking frames based on their feature space proximity and their respective positions in time.
In particular, we would like this representation to encode both feature-space and temporal proximity.
We achieve this by using time progression as a modulating factor when constructing the graph.

For a video with N frames $X=\left \{\mathbf{x_1}, \mathbf{x_2}, \cdots, \mathbf{x_N}\right \}$, we define a directed graph $G=(V,E)$ with edges describing the proximity of frames in feature space and time. We construct $G$ by computing the frames' feature space distances and then modulating them by their respective temporal positions, using the following: 

\begin{equation}
G_f(i,j) =
\begin{dcases*}
   \text{$1 - \langle\mathbf{x_i}\,, \mathbf{x_j}\rangle $}                        & if $i \neq j$\\
   1                         & otherwise
\end{dcases*} 
\label{eq:2}
\end{equation}
where $G_f$ represents a graph with edge weights computed in the feature-space. The inner product is computed on L2-normalized feature vectors to ensure that the distance is in the $[0,1]$ range. A similar graph $G_t$ is defined in the time-space, and edge weights are computed from the time-stamps. For N frames the time-stamps are defined as $T=\left \{1, 2,\cdots, N\right \}$, and the edge weights are computed as:
\begin{equation}
G_t(i,j) =
\begin{dcases*}
   \text{$\left |t_i- t_j|\right/N$}          & if $i \neq j$\\
   1                           & otherwise
\end{dcases*} 
\label{eq:3}
\end{equation}
The edges in Equation~\ref{eq:3} represent the temporal difference between the nodes, weighted by the total length of the sequence. Because we want to use the temporal graph as a modulating factor for the feature-space graph, The term $\left |t_i- t_j|\right/N$  provides a
weighing mechanism relative to the sequence length. We then compute temporally-modulated appearance-based distances as follows:
\begin{equation}
   W(i, j) = G_f(i, j) \cdot G_t(i, j).
\label{eq:4}
\end{equation}

$W(i,j)$ therefore specifies the temporally weighted distance between graph nodes ($i.e.$ frames) $i$ and $j$. Finally, from this we construct a 1-NN graph by keeping only the closest node to each node (according to $W(i,j)$) and setting all other edges to zero. 
\begin{equation}
G(i,j) =
\begin{dcases*}
   0                        & if $W(i,j) > \displaystyle\min_{\forall j} W(i,j)$ \\
   1                        & otherwise    
\end{dcases*} \\
\label{eq:5}
\end{equation}

The 1-NN temporal graph $G$ defines an adjacency matrix where each node is linked to its closest neighbor according to the temporally weighted distances $W$. For all non-zero edges $G(i,j)$, we make the links symmetric by setting $G(j,i)=1$.
This results in a symmetric sparse matrix that encodes both feature space and temporal distances, and whose connected components form clusters.
Note that Equation~\ref{eq:5} only creates absolute links in the graph and we do not perform any additional graph segmentation steps. In contrast, popular methods that build similar nearest-neighbour graphs, such as spectral clustering~\cite{von2007tutorial}, need to solve a graph-cut problem that involves solving an eigenvalue decomposition and thus have cubic complexities.

The connected components of the graph in Equation~\ref{eq:5} automatically partition the data into discovered clusters.
We use a recursive procedure to obtain further successive groupings of this partition.
Each step forms groups of previously-obtained clusters, and the recursion terminates when only one cluster remains.
Because in each recursion the graph's connected components form larger linking chains~\cite{finch}, in only a few recursions a small set of hierarchical partitions can be obtained, where each successive partition contains clusters of the previous partition's clusters.

The main steps of the proposed algorithm are shown in Algorithm~\ref{algo:the_alg}. After computing the temporal 1-NN graph through Equations~\ref{eq:2}-\ref{eq:5}, its connected components provide the first partition. We then merge these clusters recursively based on the cluster averages of features and time-stamps. Algo.~\ref{algo:the_alg} produces a hierarchy of partitions where each successive partition has fewer clusters.
To provide the required number of clusters $K$ we choose a partition in this hierarchy with the minimal number of clusters that is $\geq K$. If the selected partition has more than $K$ clusters, we refine it, one merge at a time as outlined in Algo.~\ref{algo:the_alg2}, until $K$ clusters ($i.e.$ actions) remain.

\begin{figure}
    \centering
    \includegraphics[width=1\linewidth]{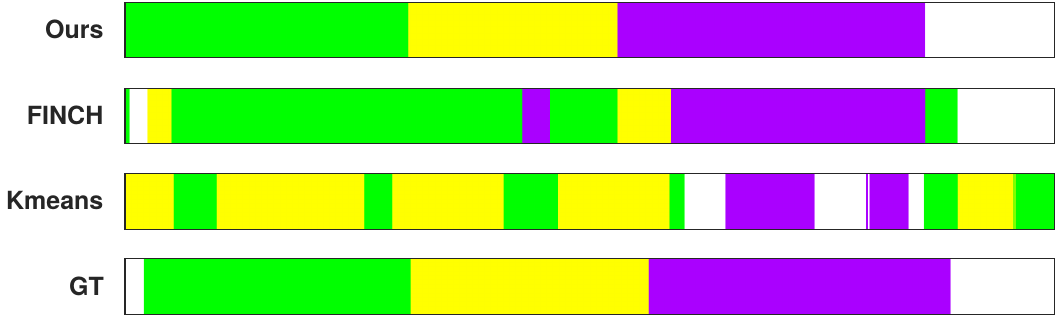}
    \caption{Segmentation output on a video from Breakfast Dataset~\cite{ute_15}: Our method provides more accurate segment lengths of actions occurring in this video.\label{fig:vis_method}}
    \vspace{-0.3cm}
\end{figure}

Note that since in each successive merge time-stamps represent the average or central time of the cluster, this automatically ensures that merged clusters are highly temporally consistent. This aspect of our proposal is important as it may provides better temporal ordering of actions. In temporal video segmentation, obtaining correct ordering of actions is crucial and quite challenging. Existing SoTA unsupervised methods~\cite{ute_paper, mallow, nnv} employ expensive post-processing mechanisms such as Generalized Mallow models~\cite{mallows1957non}, Gaussian mixture models and Viterbi decoding to improve the ordering of their predicted action segments. In contrast, because of our temporal weighing, our clustering algorithm inherently produces time-consistent clusters, thus largely preserving the correct lengths of the actions occurring in a video. In Figure~\ref{fig:vis_method} we visualize the obtained action segments and their order (by mapping the obtained segments under  Hungarian matching) on a sample video. This video depicts 4 ground-truth clusters and has $\approx800$ frames. The first step of Algo.~\ref{algo:the_alg} (lines~\ref{algo:line4}-\ref{algo:line5}) provides a partition of these $800$ frames with $254$ clusters. The successive merges of this partitioning produce 3 hierarchical partitions with $67$, $20$, $3$ and $1$ cluster(s) and the algorithm stops in only 4 steps. 
We then use Algo.~\ref{algo:the_alg2} to obtain the required number of ground-truth action segments, 4, for this video. 
The partition with the minimal number of clusters $\geq 4$ (in this case partition 3 with 20 clusters) is refined one merge at a time to produce the 4 clusters or action segments. Note that, in direct contrast to Kmeans and FINCH, our temporally-weighted clustering provides better action segments and also preserves their order in the video. 

While we use a similar procedure for hierarchical merges as in FINCH~\cite{finch}, our work differs in scope and technical approach.
In our proposal we build a temporally modulated 1-NN graph which, unlike FINCH, requires us to use all pairwise distances of samples both in space and in time for building the adjacency matrix. Our method, thus, can be considered a special case of FINCH which is well suited for videos. Because of these differences, and for clarity in comparing both, we term our method \textbf{T}emporally \textbf{W}eighted \textbf{FI}rst \textbf{N}N \textbf{C}lustering \textbf{H}ierarchy (TW-FINCH). 
For action segmentation, TW-FINCH shows clear performances advantages over both Kmeans and FINCH, as we will show next in section~\ref{sec:exp}.

\begin{algorithm}[t]
\caption{Temporally Weighted Clustering Hierarchy}\label{algo:the_alg}
\begin{algorithmic}[1]
\STATE \textbf{Input:} Video $X=\left \{\mathbf{x_1}, \mathbf{x_2}, \cdots, \mathbf{x_N}\right \}$, $X \in \mathbb{R}^{ N \times d}$ 
\STATE \textbf{Output:} Set of Partitions $\mathbf{\mathcal{S}}=\{\mathcal{P}_1,\mathcal{P}_2,\cdots,\mathcal{P}_S\}$ such that $\mathcal{P}_{i+1}\supseteq\mathcal{P}_i$ $\forall i\in \mathcal{S}$. Each partition $\mathcal{P}_i=\{C_1,C_2,\cdots,C_{\mathcal{P}_i}\}$ is a valid clustering of $X$. 
\STATE \textbf{Initialization:}
\STATE Initialize time-stamps $T=\left \{1, 2,\cdots, N\right \}$. Compute 1-NN temporally weighted graph $G$  via Equation~\ref{eq:2}-\ref{eq:5} \label{algo:line4}
\STATE  Get first partition $\mathcal{P}_1$ with $C_{\mathcal{P}_1}$ clusters from connected-components of $G$  . \label{algo:line5}
\WHILE{there are at least two clusters in $\mathcal{P}_i$}
\STATE  Given input data $X$ and its partition $\mathcal{P}_i$ prepare averaged data matrix $M=\{\bar{\mathbf{x}}_1, \bar{\mathbf{x}}_2,\cdots,\bar{\mathbf{x}}_{C_{\Gamma_i}}\}$ and averaged time-stamps $T_M=\{\bar{t}_1,\bar{t}_2,\cdots, \bar{t}_{C_{\mathcal{P}_i}}\}$ , where   $\mathbb{M}^{ C_{\mathcal{P}_i} \times d}$ and $\mathbb{T_M}^{ C_{\mathcal{P}_i} \times 1}$. 
\STATE Compute 1-NN temporally weighted graph $G_M$  via Equation~\ref{eq:2}-\ref{eq:5} with feature vectors in $M$ and time-stamps in $T_M$.
\STATE Get partition $\mathcal{P}_M$ of $\mathcal{P}_i$ from connected-components of $G_M$.
\IF {$\mathcal{P}_M$ has one cluster}
\STATE break
\ELSE
\STATE Update cluster labels in $\mathcal{P}_i:\mathcal{P}_M\to\mathcal{P}_i$
\ENDIF 
\ENDWHILE
\end{algorithmic}
\end{algorithm}

\begin{algorithm}[t]
\caption{Final Action Segmentation}\label{algo:the_alg2} 
\begin{algorithmic}[1]
\STATE \textbf{Input:} \# of actions $K$, Video $X=\left \{\mathbf{x_1}, \mathbf{x_2}, \cdots, \mathbf{x_N}\right \}$ and a partition $\mathcal{P}_i$ from the output of Algorithm~1.
\STATE \textbf{Output:} Partition $\mathcal{P}_K$ with required number of action labels.
\STATE \textbf{Merge two clusters at a time:}
\FOR{steps = \# of clusters in $\mathcal{P}_i$ - $K$}
\STATE  Initialize time-stamps $T=\left \{1, 2,\cdots, N\right \}$. Given input data $X$ and its partition $\mathcal{P}_i$ prepare averaged data matrix $M=\{\bar{\mathbf{x}}_1, \bar{\mathbf{x}}_2,\cdots,\bar{\mathbf{x}}_{C_{\mathcal{P}_i}}\}$ and averaged time-stamps $T_M=\{\bar{t}_1,\bar{t}_2,\cdots, \bar{t}_{C_{\mathcal{P}_i}}\}$
\STATE Compute 1-NN temporally weighted graph $G_M$  via Equation~\ref{eq:2}-\ref{eq:5}
\STATE  $\forall G_M(i,j)=1$ keep only one symmetric link $(i,j)$ with the minimum temporal distance $W(i,j)$ obtained in Equation~\ref{eq:4} and set all others to zero. 
\STATE  Update cluster labels in $\mathcal{P}_i$: Merge corresponding $i,j$ clusters in $\mathcal{P}_i$
\ENDFOR
\end{algorithmic}

\end{algorithm}

\section{Experiments}
\label{sec:exp}
In this section, we first introduce the datasets, features, and metrics used to evaluate our TW-FINCH method, before comparing it both to baseline and SoTA approaches.\\

\noindent\textbf{Datasets:} We conduct experiments on five challenging and popular temporal action segmentation datasets, namely Breakfast~(BF)~\cite{ute_15}, Inria Instructional Videos~(YTI)~\cite{yti_paper}, 50Salads~(FS)~\cite{fs_paper}, MPII Cooking 2~(MPII)~\cite{mpii_paper}, and Hollywood Extended~(HE)~\cite{he_paper}. As shown in Table~\ref{table:datasets}, these 5 datasets cover a wide variety of activities (from cooking different types of meals to car maintenance), contain videos of varying lengths (from 520 frames on average to up to 11788 frames), and have different levels of average action granularity (from 3 up to 19).

\begin{table}[t!]
\vspace{0.3cm}
\centering
\resizebox{8.5cm}{!}{
\begin{tabular}{l|ccccc}
\toprule
& BF~\cite{ute_15}    &   YTI~\cite{yti_paper} &   FS~\cite{fs_paper} &MPII~\cite{mpii_paper}  & HE~\cite{he_paper} \\
\midrule
\#Videos    & 1712  &   150 &   50 & 273    & 937 \\
 Avg. \#Frames-per-video     & 2099  & 520 &  11788     & 10555   & 835    \\  
Feature Dim.        & 64    &   3000&   64  &64      & 64  \\
\#Activities~(V)  & 10    &   5    &   1   & 67       & 16 \\  
Avg. \#Actions-per-video~(A)  & 6   &  9    &  19    &    17    & 3 \\  
Background  & 7\% & 63.5\%  & 14.1\% & 29\% & 61\%\\
\bottomrule
\end{tabular}}
\caption{Statistics of datasets used in the experiments: Background refers to the \% of background frames in a dataset.\label{table:datasets}}
\vspace{-0.3cm}
\end{table}

\noindent\textbf{Features:} To ensure a fair comparison to related work, we use the same input features that were used by recent methods~\cite{ute_paper,mallow,sct,cdfl,actionset}.
Specifically, for the BF, FS, MPII, and HE datasets we use the improved dense trajectory (IDT)~\cite{idt} features computed and provided by the authors of CTE~\cite{ute_paper} (for BF and FS) and SCT~\cite{sct} (for MPII and HE).
For YTI~\cite{yti_paper}, we use the features provided by the authors, which are 3000-dimensional feature vectors formed by a concatenation of HOF~\cite{hof} descriptors and features extracted from VGG16-conv5~\cite{vgg}.
For all datasets, we report performance for the full dataset, consistent with literature. 

\noindent\textbf{Metrics:} To evaluate the temporal segmentation, we require a one-to-one mapping between the predicted segments and ground-truth labels.  Following~\cite{ute_paper,mallow,cdfl,sct,actionset}, we generate such a mapping using the Hungarian algorithm then evaluate with four metrics: (i) accuracy, calculated as the mean over frames~(MoF); (ii) the F1-score;  (iii) the Jaccard index, calculated as the intersection over union~(IoU); and (iv) the midpoint hit criterion~\cite{rohrbach2012database}, where the midpoint of the predicted segment must be within the ground-truth. We report MoF and IoU for all datasets, and in addition F1-score for YTI and midpoint hit for MPII, as used in previous works. For all metrics, a higher result indicates better performance.

\begin{table}[t!]
\centering
\resizebox{8.3cm}{!}{
\begin{tabular}{ll|cc|c}
\toprule
	     &  \multicolumn{3}{c}{\color{blue}{Breakfast dataset}}   \\
\midrule
Supervision  &Method&IoU & MoF & T	  \\
\midrule
\multirow{7}{*}{\color{blue}{Fully Sup.}} 
&HOGHOF+HTK~\cite{ute_15}         &\textemdash&28.8& \ding{51}\\
&TCFPN~\cite{ute_7}              &\textemdash&52.0& \ding{51}\\
&HTK+DTF w. PCA~\cite{ute_16}     & 9.8 &56.3& \ding{51}\\
&GMM+CNN~\cite{ute_17}    &36.1 &50.7& \ding{51}\\
&RNN+HMM~\cite{kuehne2018hybrid}            &\textemdash&61.3& \ding{51}\\
&MS-TCN~\cite{mstcn}   &\textemdash& 66.3& \ding{51}\\
&SSTDA~\cite{sstda}       &\textemdash& 70.2& \ding{51}\\
\midrule
\multirow{10}{*}{\color{blue}{Weakly Sup.}}
&ECTC~\cite{ute_12}       &\textemdash &27.7& \ding{51}\\
&GMM+CNN~\cite{ute_17}    &12.9 &28.2& \ding{51}\\
&SCT~\cite{sct}   & \textemdash &30.4 & \ding{51} \\
&RNN-FC~\cite{ute_24} &\textemdash&33.3& \ding{51}\\
&RNN+HMM~\cite{kuehne2018hybrid}            &\textemdash&36.7& \ding{51}\\
&TCFPN~\cite{ute_7}      &24.2 &38.4& \ding{51}\\
&NN-Vit.~\cite{nnv}    &\textemdash&43.0& \ding{51}\\
&D3TW~\cite{d3tw}   &\textemdash& 45.7 & \ding{51} \\
&MuCon~\cite{mucon} &\textemdash& 49.7 & \ding{51}\\
&CDFL~\cite{cdfl}   &33.7& 50.2& \ding{51}  \\
\midrule
\multirow{3}{*}{\color{blue}{Unsup. Baselines}}
& {\color{red}{Equal Split}} 
& {\color{red}{21.9}} & {\color{red}{34.8}} & \ding{55}\\
& Kmeans & 23.5 & 42.7 &  \ding{55}\\
& FINCH & 28.3 & 51.9 & \ding{55}\\
\midrule
\multirow{6}{*}{\color{blue}{Unsup.}}
&Mallow*~\cite{mallow}  & \textemdash & 34.6 & \ding{51}\\
&CTE*~\cite{ute_paper} &\textemdash &41.8 & \ding{51}\\ 
&LSTM+AL~\cite{lstm_al} &\textemdash &42.9 & \ding{51}\\ 
&VTE-UNET~\cite{vt_unet} &\textemdash &52.2 & \ding{51}\\ 
&\textbf{TW-FINCH} & 42.3 & 62.7  & \ding{55}\\
\midrule
\color{blue}{Unsup.} &\textbf{TW-FINCH (K=gt/video)} & 44.1 & 63.8  & \ding{55}\\
\bottomrule
\end{tabular}}
\caption{Comparison on the Breakfast dataset~\cite{ute_15} (* denotes results with Hungarian computed over all videos of an activity together). T denotes whether the method has a training stage on target activity/videos.
\label{table:eval_bf}}
\vspace{-0.3cm}
\end{table}

\noindent\textbf{Evaluation Setup:} 
Recent methods~\cite{lstm_al, ute_paper,mallow,cdfl, vt_unet} all evaluate at ground-truth number of actions for an activity.

We adopt a similar approach and set $K$, for a video of a given activity, as the average number of actions for that activity. To provide an upper limit on the performance of our method we also evaluate with K set as the groundtruth of each video. 



\subsection{Comparison with baseline methods}
As established in section~\ref{sec:method}, Kmeans~\cite{kmeans} and FINCH~\cite{finch} are strong baselines for temporal action segmentation.
In this section we establish an additional baseline, which we call \textit{Equal Split}, that involves simply splitting the frames in a video into $K$ equal parts.
It can be viewed as a temporal clustering baseline based only on the relative time-stamps of each frame.
This seemingly trivial baseline is competitive for all datasets and actually outperforms many recent weakly-supervised and unsupervised methods for the BF~(Table~\ref{table:eval_bf}) and FS~(Table~\ref{table:eval_fs}) datasets.
TW-FINCH, however, consistently outperforms all baselines by significant margins on all five datasets, as shown in Table~\ref{table:eval_bf}~(BF), Table~\ref{table:eval_yti}~(YTI), Table~\ref{table:eval_fs}~(FS), Table~\ref{table:eval_mpii}~(MPII) and Table~\ref{table:eval_he}~(HE).
We attribute these strong results to better temporal consistency and ordering of actions, which TW-FINCH is able to achieve due to temporal weighting.

\subsection{Comparison with the State-of-the-art} We now compare TW-FINCH
to current state-of-the-arts, discussing results for each of the 5 datasets in turn. However, as noted in~\cite{ute_paper} even though evaluation metrics are comparable to weakly and fully
supervised approaches, one needs to consider that the results of the unsupervised learning are reported with respect
to an optimal assignment of clusters to ground-truth classes
and therefore report the best possible scenario for the task.
For each dataset, we report IoU and MoF results for TW-FINCH. We report additional metrics when they are commonly used for a given dataset.

The column \textbf{T} in the tables denotes whether the method requires training on the target videos of an activity before being able to segment them. A dash indicates no known reported results.\\
%
%

\noindent\textbf{Breakfast dataset~(BF):} BF contains an average of 6 actions per video, and 7\% of frames in the dataset are background frames.

In Table~\ref{table:eval_bf} we report results on BF and compare TW-FINCH with recent state-of-the-art unsupervised, weakly-supervised and fully-supervised approaches. TW-FINCH outperforms all unsupervised methods, with absolute improvements of
$10.5$\% over the best reported unsupervised method VTE-UNET and $19.8$\% over LSTM+AL~\cite{lstm_al}. Similarly TW-FINCH outperforms the best reported weakly-supervised method CDFL~\cite{cdfl} with a $8.6/12.5$\% gain on the IoU/MoF metrics.

Methods \cite{ute_paper, mallow} train a separate segmentation model for each activity, and set $K$ to the maximum number of groundtruth actions for that activity. They then report results by computing Hungarian over all videos of one activity. Since we are clustering each video separately, using $K$ as maximum would over segment most of the videos. This however still enable us to show the impact on performance in such a case.
When we set K to the maximum \# actions per activity on the BF dataset, our performance is $57.8\%$, as many of the videos are over segmented. 
To see the purity of these over-segmented clusters we computed the weighted cluster purity in this setting, which comes out to be $83.8\%$.
This high purity indicates that, even with an inexact $K$, our clusters can still be used for downstream tasks such as training self-supervised video recognition models.

\begin{table}[t!]
\centering
\resizebox{8.3cm}{!}{
\begin{tabular}{ll|cc|c}
\toprule
	     &  \multicolumn{3}{c}{\color{blue}{Inria Instructional Videos}}   \\
\midrule
Supervision  &Method&F1-Score & MoF & T	  \\
\midrule
\multirow{3}{*}{\color{blue}{Unsup. Baselines}}
& {\color{red}{Equal Split}} 
& {\color{red}{27.8}} & {\color{red}{30.2}} & \ding{55}\\
& Kmeans & 29.4 & 38.5 & \ding{55}\\
& FINCH & 35.4 & 44.8 & \ding{55}\\
\midrule
\multirow{7}{*}{\color{blue}{Unsup.}}
&Mallow*~\cite{mallow}  &27.0 &27.8 &\ding{51} \\
&CTE*~\cite{ute_paper} &28.3 &39.0 & \ding{51} \\
&LSTM+AL~\cite{lstm_al} &39.7 &\textemdash & \ding{51} \\
&\textbf{TW-FINCH} & 48.2 & 56.7 & \ding{55}\\
\midrule
\color{blue}{Unsup.} &\textbf{TW-FINCH (K=gt/video)} & 51.9 & 58.6 & \ding{55}\\
\bottomrule
\end{tabular}}
\caption{Comparison on the Inria Instructional Videos~\cite{yti_paper} dataset. * denotes results with Hungarian computed over all videos of an activity together.
\label{table:eval_yti}}
\end{table}

%
%

\noindent\textbf{Inria Instructional Videos~(YTI):}\ YTI contains an average of 9 actions per video, and 63.5\% of all frames in the dataset are background frames.

In Table~\ref{table:eval_yti} we summarize the performance of TW-FINCH on YTI and compare to recent state-of-the-art unsupervised and weakly-supervised approaches. To enable direct comparison, we follow previous works and remove a ratio~($\tau=75\%$) of the background frames from the video sequence and report the performance. TW-FINCH  outperforms other methods and achieves F1-Score of 48.2\% and MoF of 56.7\%, which constitute absolute improvements of 8.5\% on F1-Score over the best published unsupervised method. 

\textbf{Impact of Background on YTI.} As  63.5\% of all frames in the YTI dataset are background, methods that train on this dataset tend to over-fit on the background. 
In contrast, a clustering based method is not strongly impacted by this: when we evaluate TW-FINCH while including all of the background frames our MoF accuracy drops from 56.7 $\rightarrow$ 43.4\% as is expected due to having more frames and thus more errors. Given such a significant data bias on background frames this relatively small drop indicates 
that TW-FINCH works reasonably with widely-varying degrees of background content.

\begin{table}[t!]
\centering
\resizebox{8.3cm}{!}{
\begin{tabular}{ll|cc|c}
\toprule
	     &  \multicolumn{3}{c}{\color{blue}{50Salads}}   \\
\midrule
Supervision  &Method&eval & mid & T	  \\
\midrule
\multirow{5}{*}{\color{blue}{Fully Sup.}}
& ST-CNN~\cite{stcnn} &68.0 & 58.1& \ding{51}\\
& ED-TCN~\cite{edtcn}   &72.0 & 64.7& \ding{51}\\
& TricorNet~\cite{ute_6} &73.4 &67.5& \ding{51}\\
& MS-TCN~\cite{mstcn} &80.7 &\textemdash & \ding{51}\\
& SSTDA~\cite{sstda} & 83.8 &\textemdash  & \ding{51}\\
\midrule
\multirow{5}{*}{\color{blue}{Weakly Sup.}}
& ECTC~\cite{ute_12}    & \textemdash & 11.9& \ding{51}\\
& HTK+DTF~\cite{ute_16} & \textemdash &24.7& \ding{51}\\
& RNN-FC~\cite{ute_24}  & \textemdash &45.5& \ding{51}\\
& NN-Vit.~\cite{nnv}    & \textemdash &49.4& \ding{51}\\
& CDFL~\cite{cdfl}      & \textemdash &54.7& \ding{51}\\
\midrule
\multirow{3}{*}{\color{blue}{Unsup. Baselines}}
& {\color{red}{Equal Split}} 
& {\color{red}{47.4}} & {\color{red}{33.1}}  & \ding{55}\\
& Kmeans & 34.4 & 29.4  & \ding{55}\\
& FINCH & 39.6 & 33.7 & \ding{55}\\
\midrule
\multirow{3}{*}{\color{blue}{Unsup.}}
&LSTM+AL~\cite{lstm_al}  &60.6      &\textemdash & \ding{51} \\
&\textbf{TW-FINCH} & 71.1 & 66.5 & \ding{55}\\
\midrule
\color{blue}{Unsup.} &\textbf{TW-FINCH (K=gt/video)} & 71.7 & 66.8 & \ding{55}\\
\bottomrule
\end{tabular}}
\caption{Comparison to SoTA approaches at \textit{eval} and \textit{mid} granularity levels on the 50Salads dataset~\cite{fs_paper}. We report \textbf{MoF}. 
\label{table:eval_fs}}
\end{table}

%
%

\noindent\textbf{50Salads~(FS):} FS contains an average of 19 actions per video, and 14.1\% of all frames in the dataset are background frames.
We evaluate with respect to two action granularity levels, as described in~\cite{fs_paper}. The \textbf{\textit{mid}} granularity level evaluates performance on the full set of 19 actions while the \textbf{\textit{eval}} granularity level merges some of these action classes, resulting in 10 action classes. In Table~\ref{table:eval_fs} we show that TW-FINCH obtains a MoF of 66.5\% in the \textit{mid} granularity, 11.8\% higher (in absolute terms) than the best weakly-supervised method CDFL~\cite{cdfl}. 
We see similar performance gains in the \textit{eval} granularity level evaluation as well. The IoU score of TW-FINCH for \textit{mid} and \textit{eval} granularity is 48.4\% and 51.5\% respectively.
%
%

\begin{table}[t]
\centering
\resizebox{8.3cm}{!}{
\begin{tabular}{ll|cccc|c}
\toprule
	     &  \multicolumn{5}{c}{\color{blue}{MPI Cooking 2}}   \\
\midrule
Supervision  &Method&IoU &\multicolumn{2}{c}{{Midpoint-hit}}&MoF	&T  \\
	        && &Precision &Recall&	&  \\
\midrule
\multirow{3}{*}{\color{blue}{Fully Sup.}}
& Pose + Holistic~\cite{rohrbach2012database} &\textemdash &19.8& 40.2& \textemdash& \ding{51}\\
& Fine-grained~\cite{ni2014multiple} &\textemdash &28.6& 54.3& \textemdash& \ding{51}\\
& GMM+CNN~\cite{ute_17} & 45.5 &\textemdash& \textemdash& 72.0& \ding{51}\\
\midrule
\multirow{1}{*}{\color{blue}{Weakly Sup.}}
& GMM+CNN~\cite{ute_17} & 29.7 &\textemdash& \textemdash& 59.7& \ding{51}\\

\midrule
\multirow{3}{*}{\color{blue}{Unsup. Baselines}}
& {\color{red}{Equal Split}}
& {\color{red}{6.9}} &{\color{red}{25.6}}& {\color{red}{44.6}} & {\color{red}{14.6}} & \ding{55} \\
& Kmeans & 14.5 &21.9& 34.8 & 30.4&  \ding{55} \\
& FINCH & 18.3 &26.3& 41.9 &  40.5 & \ding{55}\\
\midrule
\multirow{1}{*}{\color{blue}{Unsup.}}
 & \textbf{TW-FINCH} & 23.1 & 34.1 &54.9  & 42.0 & \ding{55} \\
\midrule
\color{blue}{Unsup.} &\textbf{TW-FINCH (K=gt/video)} & 24.6 &37.5& 59.2 & 43.4 & \ding{55} \\

\bottomrule
\end{tabular}}
\caption{Comparison on the MPII Cooking 2 dataset~\cite{mpii_paper}.
\label{table:eval_mpii}}
\end{table}

\noindent\textbf{MPII Cooking 2~(MPII):} MPII contains 17 actions per video on average, and 29\% of all frames in the dataset are background frames. For MPII we report the midpoint hit criterion~\cite{rohrbach2012database} (multi-class precision and recall), the standard metric for this dataset, in addition to IoU and MoF. The dataset provides a fixed train/test split. We report performance on the test set to enable direct comparisons with previously reported results.
As Table~\ref{table:eval_mpii} shows, TW-FINCH outperforms our strong unsupervised baselines for all 4 reported metrics. Our method also outperforms SoTA fully-supervised methods that report the mid-point hit criterion.

%
%

\begin{table}[t]
\centering
\resizebox{8.3cm}{!}{
\begin{tabular}{ll|cc|c}
\toprule
&  \multicolumn{3}{c}{\color{blue}{Hollywood Extended}}   \\
\midrule
Supervision  &Method&IoU & MoF &T	  \\
\midrule
\multirow{1}{*}{\color{blue}{Fully Sup.}}
& GMM+CNN~\cite{ute_17} &  8.4& 39.5 &\ding{51}\\
\midrule
\multirow{7}{*}{\color{blue}{Weakly Sup.}}
& GMM+CNN~\cite{ute_17}   & 8.6 & 33.0 & \ding{51}\\
& ActionSet~\cite{actionset}&9.3  &\textemdash& \ding{51} \\
& RNN-FC~\cite{ute_24}  & 11.9 &\textemdash & \ding{51}\\
& TCFPN~\cite{ute_7}    & 12.6 & 28.7 & \ding{51}\\
& SCT~\cite{sct}        & 17.7 & \textemdash& \ding{51} \\
& D3TW~\cite{d3tw}      & \textemdash & 33.6& \ding{51} \\
& CDFL~\cite{cdfl}      & 19.5 & 45.0 & \ding{51} \\
\midrule 
\multirow{3}{*}{\color{blue}{Unsup. Baselines}}
& {\color{red}{Equal Split}} 
& {\color{red}{24.6}} & {\color{red}{39.6}} & \ding{55}\\
& Kmeans & 33.2 & 55.3 & \ding{55}\\
& FINCH & 37.1 & 56.8 & \ding{55}\\
\midrule
\multirow{1}{*}{\color{blue}{Unsup.}}
& \textbf{TW-FINCH} & 35.0  & 55.0 & \ding{55} \\
\midrule
\color{blue}{Unsup.} &\textbf{TW-FINCH (K=gt/video)} & 38.5  & 57.8 & \ding{55} \\
\bottomrule
\end{tabular}}
\caption{Comparison on the Hollywood Extended dataset~\cite{he_paper}. 
\label{table:eval_he}}
\end{table}

\noindent\textbf{Hollywood Extended~(HE):} HE contains an average of 3 (including background) actions per video, and 61\% of all frames in the dataset are background frames.  
We report results for HE in Table~\ref{table:eval_he}, which shows that TW-FINCH outperforms CDFL~\cite{cdfl} by 15.5\% (19.5$\rightarrow$35.0) and 10.0\% (45$\rightarrow$55.0) in IoU and MoF, respectively. Further, note that the performance of our appearance-based clustering baselines is quite similar to the performance of our method. We attribute this to the small number of clusters per video (3 clusters on average). As a result, Kmeans and FINCH are roughly as effective as TW-FINCH, as temporally ordering 3 clusters is less difficult.\\

\begin{figure*}[t]
\centering
    \begin{minipage}[b]{0.9\columnwidth}
    \centering
    \centerline{\includegraphics[width=1\columnwidth]{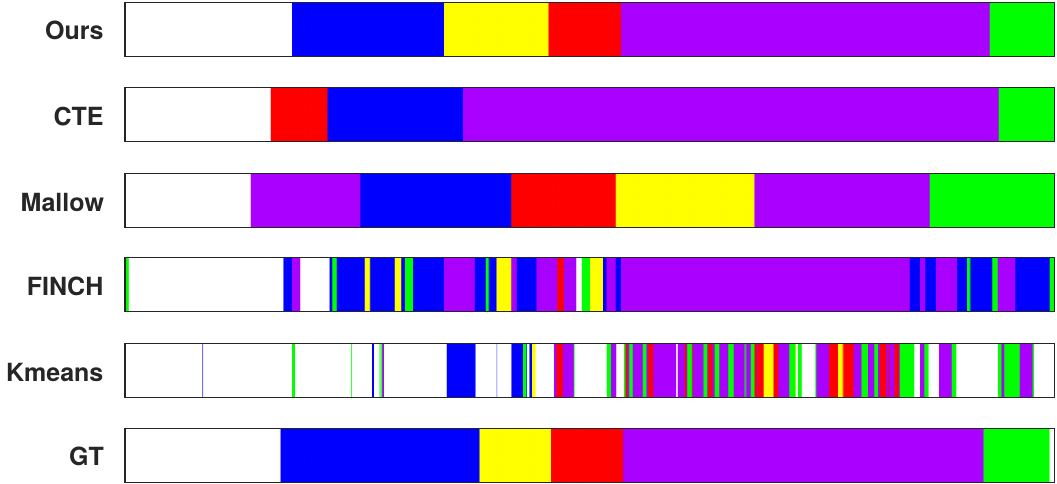}\vspace{0.1cm}}
    \centerline{(a) \textit{P39\_cam02\_P39\_scrambledegg}}
    \end{minipage} \quad
    \begin{minipage}[b]{0.9\columnwidth}
    \centering
    \centerline{\includegraphics[width=1\columnwidth]{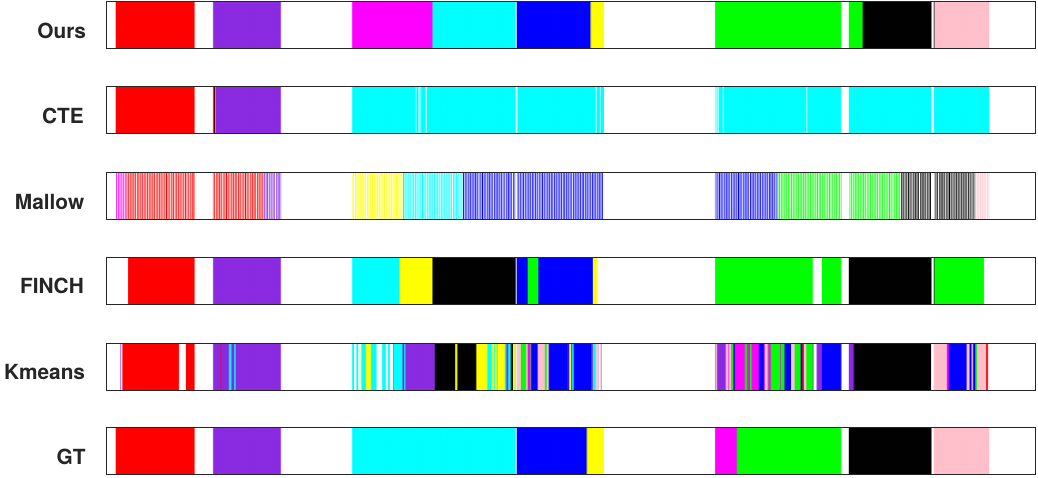}\vspace{0.1cm}}
    \centerline{(b) \textit{changing\_tire\_0023}
    
    }
    \end{minipage} 
    \caption{Segmentation examples from (a) the Breakfast dataset~\cite{ute_15}, and (b) the Inria  Instructional Videos dataset~\cite{yti_paper}. Colors indicate different actions and are arranged in chronological order. We compare the segmentation quality of our method to Kmeans, FINCH, and two state-of-the-art unsupervised methods, CTE and Mallow. Our method has predicted better lengths of actions occurring in these videos.\label{fig:qual_ac}}
\end{figure*}

\noindent\textbf{Qualitative Results.} Fig.~\ref{fig:qual_ac} shows representative results for two videos taken from the BF dataset (a) and the YTI dataset (b).
Note that in the visualization (b) we set the background frames to white for all methods, according to the ground-truth. This allows the misclassification and mis-ordering of background frames to be more clearly seen.
In (a), one can observe that TW-FINCH accurately predicts the length of segments, yielding better segmentation boundaries. Both clustering baselines, neither of which leverage temporal information, have noisy segments. Other SoTA unsupervised methods either have inaccurate temporal boundaries, incorrect ordering of actions, or are missing actions altogether. In addition, background frames are more often misclassified and mis-ordered for competing methods. Similar observations can be made in (b), where we show qualitative segmentation results on a more challenging YTI video with 9 actions of varying lengths, and several interspersed background scenes.\\

\noindent\textbf{Limitations.} Two main limitations for our work exist, which are inherent to our unsupervised clustering-based approach. The first, illustrated in Figure~\ref{fig:qual_ac}(b), occurs when we over-segment a temporally-contiguous sequence due to low visual coherence. 
The second may occur when we assign frames that depict the same action to different clusters because they are temporally distant.\\

\noindent\textbf{Computational complexity.} As we need to compute the $N \times N$ temporal distances, the computational complexity of TW-FINCH is $\mathcal{O}(N^2)$. In contrast FINCH is $\mathcal{O}(Nlog(N))$ while other similar graph-based clustering methods such as spectral methods are $\mathcal{O}(N^3)$ and hierarchical agglomerative linkage-based schemes are $\mathcal{O}(N^2log(N))$.
Table~\ref{table:time} provides the total run-time of TW-FINCH and other state-of-the-art methods on Breakfast dataset split 1 (252 videos). Unlike previous methods that require hours of model training on GPUs, our method runs on a computer with an AMD 16-core processor, taking approximately $0.16$ seconds on average to segment one video ($\approx2000$ frames).

\begin{table}[t!]
\centering
\resizebox{8.3cm}{!}{
\begin{tabular}{ll|cc|c}
\toprule
Supervision &Method & Training & Testing &T \\
&      & (hours) & (seconds) \\
\midrule
\multirow{4}{*}{\color{blue}{Weakly Sup.}}
&TCFPN$^{\star}$~\cite{ute_7} & 12.75 & 00.01& \ding{51}\\
&NN-Vit.$^{\star}$~\cite{nnv} & 11.23 & 56.25 & \ding{51}\\
&CDFL$^{\star}$~\cite{cdfl}   & 66.73 & 62.37 & \ding{51}\\
&MuCon-full$^{\star}$~\cite{mucon} & 04.57 & 03.03 & \ding{51}\\
\midrule
\multirow{2}{*}{\color{blue}{Unsup. Baselines}} 
& Kmeans & {\color{red}{00.00}} & 38.69 & \ding{55}\\
& FINCH &  {\color{red}{00.00}} & 37.08 & \ding{55}\\
\midrule
\multirow{2}{*}{\color{blue}{Unsup.}} 
& CTE~\cite{ute_paper} & \textemdash & 217.94 & \ding{51}\\
& \textbf{TW-FINCH (Ours)} & {\color{red}{00.00}} & 40.31 & \ding{55}  \\
\bottomrule
\end{tabular} }
\caption{ Run-time comparison of method with other state-of-the-art methods on Breakfast dataset. 
Testing duration is measured as the average inference for split 1 test
set (252 videos).  $^{\star}$The run-time of all the {\color{blue}{Weakly Sup.}} methods were taken from~\cite{mucon}   \label{table:time}}
\vspace{-0.3cm}

\end{table}

\section{Conclusion}\label{sec:conclusion}
We addressed the problem of temporal action segmentation and found that simple clustering baselines produce results that are competitive with, and often outperform, recent SoTA unsupervised methods. We then proposed a new unsupervised method, TW-FINCH which encodes spatiotemporal similarities between frames on a 1-nearest-neighbor graph and produces a hierarchical clustering of frames. Our proposal is practical as unlike existing approaches it does not require training on the target activity videos to produce its action segments. Our extensive quantitative experiments demonstrate that TW-FINCH is effective and consistently outperforms SoTA methods on 5 benchmark datasets by wide margins on multiple metrics.


{\small
\bibliographystyle{ieee_fullname}
\bibliography{egbib}
}

\end{document}